\documentclass{article}
\usepackage{spconf,amsmath,graphicx}
\usepackage{amssymb}

\title{VLSTM: Very Long Short-Term Memory Networks for High-Frequency Trading}
\twoauthors
 {Prakhar Ganesh\sthanks{The author performed the work during internship at WealthNet Advisors}}
	{Research Engineer\\
	Advanced Digital Sciences Center, Singapore}
    {Puneet Rakheja}
	{Founder and CEO\\
	WealthNet Advisors, New Delhi}

\begin{document}
%
\maketitle
\begin{abstract}
Financial trading is at the forefront of time-series analysis, and has grown hand-in-hand with it. The advent of electronic trading has allowed complex machine learning solutions to enter the field of financial trading. Financial markets have both long term and short term signals and thus a good predictive model in financial trading should be able to incorporate them together. One of the most sought after forms of electronic trading is high-frequency trading (HFT), typically known for microsecond sensitive changes, which results in a tremendous amount of data. LSTMs are one of the most capable variants of the RNN family that can handle long-term dependencies, but even they are not equipped to handle such long sequences of the order of thousands of data points like in HFT. We propose very-long short term memory networks, or VLSTMs, to deal with such extreme length sequences. We explore the importance of VLSTMs in the context of HFT. We compare our model on publicly available dataset and got a 3.14\% increase in F1-score over the existing state-of-the-art time-series forecasting models. We also show that our model has great parallelization potential, which is essential for practical purposes when trading on such markets.

\end{abstract}

\section{Introduction}
\label{sec:introduction}

The usage of predictive models to infer future prices of various commodities using historical data is not new in quantitative finance, and is commonly referred to as algorithmic trading \cite{algo_trade, algo_trade2}. Older methods of algorithmic trading includes mathematical modeling of the data, for example CAPM, Fama and French factors \cite{RePEc:eee:jfinec:v:33:y:1993:i:1:p:3-5}. However, these models require precise modeling and are not capable of handling the noisy and irrational behaviour of the financial markets.

Deep learning methods are prophesied to revolutionize the field of machine learning (ML) and represent a step towards building autonomous systems. These methods are generally more robust to noises and can learn to model based on the input data. Various advancements have been made in the field of time-series analysis using deep learning, which includes simpler models like LSTMs \cite{hochreiter1997long} and RNNs, as well as more complicated additions like attention mechanism \cite{bahdanau2014neural}. While these methods have gained success in the world of time-series analysis, the requirements of algorithmic trading are very unique.

High Frequency Trading (HFT) is one of the extreme forms of electronic trading. The special challenges for ML presented by HFT can be considered two fold : (i) Microsecond sensitive live trading - As the complexity of the model increases, it gets more computationally expensive to keep up with the speed of live trading and actually use the information provided by the model, and (ii) Tremendous amount and fine granularity of data - The historical data available in HFT is extremely lengthy yet precise. However there is a lack of understanding of how the low-level data (changing every microsecond) and the high-level data (changing every few minutes) can together relate to actionable circumstances \cite{Kearns_machinelearning}.

In this paper, we propose VLSTMs, a novel variant of LSTMs, which are capable of handling time-series sequences with length of the order of thousands. We achieve this by dealing with the input sequence at multiple frequencies, allowing us to separately process both low-level and high-level signals and then combine the information together for better final performance. Since different levels of the signal are processed in parallel, this makes it easy to do them simultaneously. Thus our model does not compromise its execution speed, even though the computational requirements of our model has increased. Our contributions include :
\begin{itemize}
    \setlength\itemsep{0em}
    \item A novel LSTM variant to efficiently combine long term and short term signals from the input sequence.
    \item Improved scope of parallelization, as the model complexity is increased along the width and not the depth.
    \item Tested on publicly available datasets and found 12.37\% increase in F1 score over vanilla LSTMs and 3.14\% increase over current state-of-the-art.
\end{itemize}

The rest of the paper is organized as follows. Section \ref{sec:background} introduces financial background and reviews related work in this field. Section \ref{sec:solution} introduces our VLSTM architecture. Section \ref{sec:evaluation} presents extensive evaluation results to support our model design and Section \ref{sec:conclusion} concludes the discussion.

\section{Background}
\label{sec:background}

\subsection{Tick Data and Bid-Ask Spread}
Tick data is the raw, uncompressed data of the trading behavior available from electronically traded markets. Every order request, change in the state of order book and trade information is registered as a "tick" event.
Bid and Ask are the prices that buyers and sellers are willing to transact at, the bid for buying, and the ask for selling. A transaction occurs when either a potential buyer is willing to pay the ask price, or a potential seller is willing to accept the bid price. The mid-price is the mean price of the top bid and top ask price \cite{bid_ask}.

\subsection{Mean Reversion}
Mean reversion is a financial theory which suggests that the price of a stock tends to return towards its long running mean price over time \cite{bid_ask} and such a behavior is seen in most of the stock markets across the world \cite{Chakraborty2011MarketMA}. Trading on this strategy is done by noticing companies whose stock values have significantly moved away in some direction from its long running mean and thus is now expected to move in the opposite direction. 
Using mean reversion in stock price prediction involves both identifying the trading range for a stock (short-term information) and the evolution of the mean around which the prices will be oscillating (long-term information). 


\subsection{Related Work}

The success of deep learning models has penetrated a lot of fields, including finance. However its reach in HFT is limited \cite{dnns}, primarily due to the computational constraints and primitive problem modeling methods. While there has been some work done on the algorithmic side \cite{book_algo}, most of the work has been focused on feature engineering in HFT \cite{Kearns_machinelearning, 1709.01268, doi:10.1111/j.1540-6261.1991.tb02683.x,1011.6402, imbalance}, with simpler models like linear regression, multiple kernel learning, maximum margin etc. \cite{Kearns_machinelearning,1705.03233}

Long short-term memory networks (LSTMs) \cite{hochreiter1997long} are one of the most commonly used deep learning models for time-series analysis. Multiple variations of LSTM have been proposed over time which deal with the multi-context and long sequence length problems. For example, Hierarchical LSTMs \cite{hlstm, rehlstm} were proposed to make character level LSTMs in NLP feasible. In this architecture, the input to the upper LSTM is the output provided by the lower LSTM, thus forming a hierarchical structure. Similar networks of multi-scale LSTMs have also been proposed for applications in document modelling \cite{liu2015multi}, wind speed forecasting \cite{wind}, sound event detection \cite{lu2018multi}, etc. These kind of architectures also provide better connections for back-propagation during training on longer sequences. There have also been other additions to the vanilla LSTM, like the attention mechanism \cite{bahdanau2014neural}, which improves its ability to handle longer sequences by allowing it to attend to only a part of the sequence at a time.

\section{VLSTM}
\label{sec:solution}

\begin{figure*}
\centering
\includegraphics[width = 0.95\textwidth]{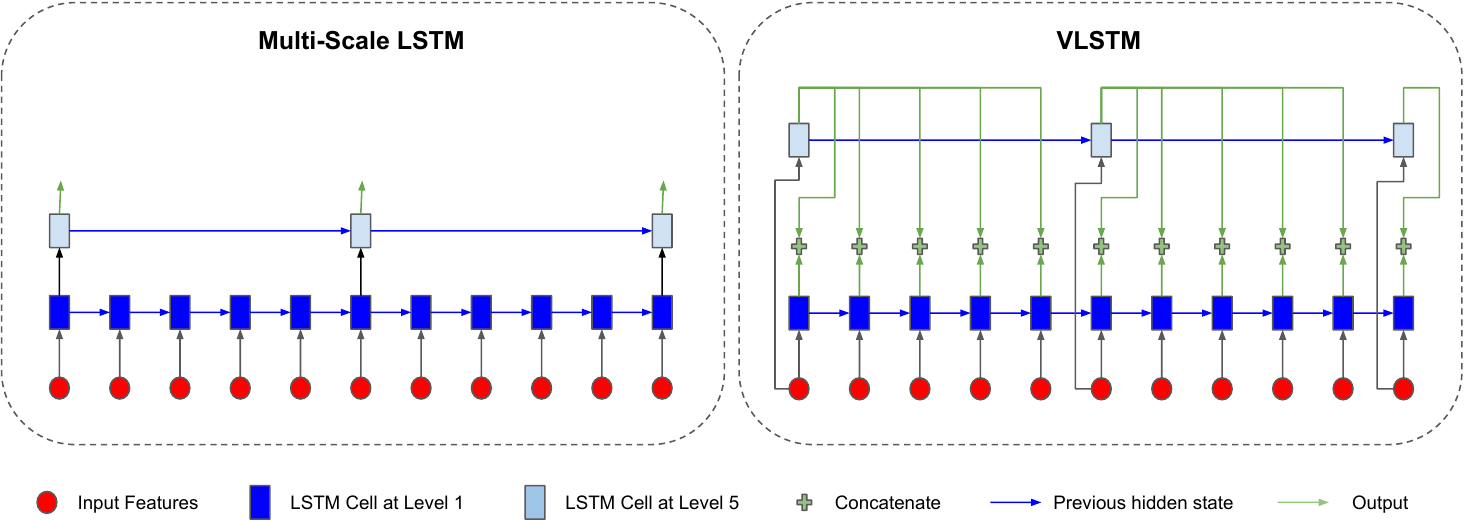}
\caption[]{Multi-Scale LSTM \{1, 5\} (left) vs VLSTM \{1, 5\} (right).}
\label{hierarvsvlstm}
\end{figure*}

We propose a novel variant of LSTM, which we call VLSTM, specialised to handle longer sequences. We first explain the theoretical motivation behind our model design and then introduce the architecture of a VLSTM.

\subsection{Motivation}

Model depth and width are two ways in which the complexity of a deep learning model can be increased. Recent advances have shown that analysing the input at varying context, which can be done by increasing the model width, can help us create a more accurate and robust model \cite{szegedy2015going}. Thus, we aim to create a deep learning model that can provide multi-context features at every input time step. 
This can be achieved by using multiple LSTM models simultaneously, each focused on extracting features from a different context level.

LSTMs working at different levels (or context) can capture a variety of information. For example, a lower-level LSTM can focus on the fine-grained and most recent behaviour of the market while a higher-level LSTM can focus on the long-term changes in the trend. However, existing variations of multi-scale LSTM models use the output provided by lower-level LSTM as input to the higher-level LSTM. This means that the higher-level LSTM do not get access to the raw input sequence and is forced to work with the lower-level LSTM for feature creation.
We hypothesize that due to this dependency, the lower-level LSTM is now focused on creating an input for the higher-level LSTM instead of creating a good feature representation of fine grained features. Similarly, the higher-level LSTM is now forced to create the final feature representation of the sequence, instead of just focusing on long-term trends present in the sequence.

We propose to decouple the functionality of LSTMs at various levels, thus allowing them to focus on only individual contexts. This can be achieved by running different LSTM levels on the raw input sequence independently, and combining the final features obtained from each level. Such a design also allows us to create a model which can provide output for every input time step, instead of only providing output at the frequency of the highest level LSTM, since at every time step atleast one of the LSTMs will update their features, thus updating the complete feature representation.

\textit{Note} : Since we are working with parallel LSTMs now, the terms lower-level and higher-level LSTM as used in multi-scale LSTMs have been adopted accordingly. Lower-level refers to LSTMs that work on the input signal at a higher frequency and are expected to provide short-term and more fine grained information, and vice-versa for higher-level LSTMs.

\subsection{Model Design}

Let the input signal be $\{x_0, x_1, x_2 .... x_n\}$, where $x_i$ represents the input vector at timestep $i$, and let there be $k$ LSTMs working at levels $\{l_1, l_2 .... l_k\}$.
For the $p^{th}$ LSTM working at level $l_p$, denote the input, output, input cell state, and output cell state respectively for the $i^{th}$ timestep as $\{W^p_i, V^p_i, C^p_{in_{i}}, C^p_{out_{i}}\}$. Based on the functioning of a conventional LSTM cell, generally we have $\{V, C_{out}\} = \sigma(W, C_{in})$, where $\sigma$ is the LSTM network.

The input to an LSTM at level $l_p$ is sampled at a frequency of $1/l_p$ from the original input sequence and the output obtained is copied for $l_p$ timesteps, i.e. until the next output is obtained. Thus, the output $Y^p_i$ obtained from the LSTM at level $l_p$ at timestep $i$ can be defined as,

\begin{equation}
\label{eq:sampling}
    Y^p_i = V^p_{\lfloor i/l_p \rfloor},
\end{equation}

\begin{equation}
\label{eq:lstm}
    \{V^p_j, C^p_{out_j}\} = \sigma^p(W^p_j, C^p_{out_{j-1}}), \;\;  W^p_j = x_{j*l_p},
\end{equation}

Once we have feature outputs from every LSTM level, these features are concatenated and passed through a dense layer followed by softmax to produce final classification outputs. We can define the final output $Y_i$ at timestep $i$ as,

\begin{equation}
\label{eq:dense}
    Y_i = \phi_C(D_C(Y^1_i \oplus Y^2_i \oplus Y^3_i, .. , Y^k_i)),
\end{equation}

where $\oplus$ denotes feature concatenation, $D_C$ denoted the dense layer for classification and $\phi_C$ denoted the final softmax activation.
Thus we get outputs \begin{math}\{Y_0, Y_1, Y_2 .... Y_n\}\end{math} which can be interpreted for predicted price movement. Refer to Fig \ref{hierarvsvlstm} for differences between a multi-scale LSTM and VLSTM.

\section{Evaluation}
\label{sec:evaluation}

We use a publicly available limit order book data to test our model and some other baseline methods. First, we do an internal ablation study of various levels in VLSTM to better understand the nature of the dataset as well as our model architecture. Next, we compare our model with existing state-of-the-art deep learning based time-series forecasting methods.

\subsection{Dataset}

We test our model on a limit order book data by doing mid-price movement prediction \cite{ntakaris2018benchmark}. Instead of using only the raw bid and ask values, several complicated hand-crafted features have also been provided by the authors, which are calculated using both static order book information as well as dynamic change in features across time. A total of 144 features are used to represent each tick \cite{ntakaris2018benchmark}. We used the publicly available version of the dataset \footnote{https://etsin.fairdata.fi/dataset/73eb48d7-4dbc-4a10-a52a-da745b47a649}.
The dataset contains 10 days of data with a total of almost 400 thousand ticks. We use the z-score based standardization provided in the dataset for our experiments. We also use the label creation method as proposed by \cite{tsantekidis2017using} in order to create 3 classes, i.e., upward and downward movement of mid-price which crosses the threshold, and movememnt of mid-price within the threshold. We use the threshold of 10\% change in mid-price in order to maintain balance between data points attributed to the 3 classes.

\subsection{Model and Experiment Settings}

We experiment with the following LSTM levels \{1, 5, 20, 100\}. Each level has 2 LSTM cells stacked on top of each other with 64 neurons each. The final dense layer used for concatenated features contains 128 neurons. Activation functions used are ReLU at every hidden layer except the classification layer. The error function used is Categorical Cross entropy with Adam optimizer.

Out of the 10 days of data present in the dataset, we use the first 6 days for training, next 1 day for validation and the final 3 days for testing. We use sliding window method to augment the training dataset and use windows of length 5,000 ticks in order to generate enough inputs for the highest-level LSTM ($l_p=100$). We allow the model to run continuously on the test and validation datasets and use mean F1 score across all 3 classes to measure their performance. We also do not make prediction for the first 5,000 ticks of the day, for the same reasons as mentioned above. We use 3 different prediction horizons $h=\{100, 200, 500\}$ (measured in number of ticks) to compare models under varying prediction targets.

\subsection{Baseline Methods}

We use 4 different models as baseline methods. First, we use a simple fully-connected multi-layer perceptron (MLP), with 3 hidden layers containing 64, 64 and 128 neurons respectively. Second, we use a vanilla LSTM model, with 2 LSTM cells of 64 neurons each stacked on top of each other, followed by a dense layer with 128 neurons. Next, we add the attention proposed by \cite{bahdanau2014neural} on top of our vanilla LSTM model. Finally, we use a multi-scale LSTM model similar to one proposed in \cite{liu2015multi}, with the same multi-level settings as our VLSTM.

\subsection{Individual LSTMs and VLSTM}

We first experimented with VLSTMs working at only one level at a time. A VLSTM working at only level 'i' is simply an LSTM working on the input signal collected with frequency '\begin{math}1/i\end{math}'. Since LSTM at every level needs to provide an output to every input timestep, the output of an LSTM working at level 'i' remains the same for 'i' ticks until the next prediction is made (as detailed in Section \ref{sec:solution}). Results for these experiments are collected in Table \ref{tab:individual}.

LSTMs working at lower levels have the advantage of changing their output more frequently but cannot process long-term information well enough to make accurate predictions. On the other hand, LSTMs working at higher levels have a lot of information on how the market has been evolving for the past few hundred ticks (or even more), but their frequency of changing outputs is extremely low.
Thus both extremes can cause a drop in F1 scores and makes the medium level LSTMs the best performing individual LSTMs. It can be noticed from the table that VLSTM at level 20 seems to outperform all other individual level VLSTMs. Being able to identify the best working individual VLSTM can also give us more insights into the market's inherent periodicity.

\begin{table}[h]
\centering
\begin{tabular}{|c|c|c|c|}
\hline
\textbf{Model $(l_1, l_2, .. l_p)$} & \multicolumn{3}{|c|}{\textbf{Mean F1}} \\
\hline
 & \textbf{h=100} & \textbf{h=200} & \textbf{h=500} \\ 
\hline
VLSTM \{1\} & 57.33\% & 55.27\% & 51.63\% \\
\hline
VLSTM \{5\} & 58.97\% & 57.86\% & 55.29\% \\
\hline
VLSTM \{20\} & \emph{63.36\%} & \emph{61.74\%} & \emph{58.78\%} \\
\hline
VLSTM \{100\} & 47.38\% & 46.69\% & 46.21\% \\
\hline
\hline
VLSTM \{1, 5, 20, 100\} & \textbf{69.70\%} & \textbf{68.41\%} & \textbf{65.90\%} \\
\hline
\end{tabular}
\caption{Performance of LSTM at various levels}
\label{tab:individual}
\end{table}

Next, we combine all the individual LSTM levels together to form our proposed VLSTM model. It can be seen that VLSTM outperforms the best performing individual LSTM (level 20) by 6.34\% and the vanilla LSTM (level 1) by 12.37\%.
This shows that adding more levels increases the variety of contextual features available to the model, which in turns improves its F1 score.

\subsection{Comparing against Baselines}

We compared our model performance against various time-series prediction baseline methods and collected the results in Table \ref{tab:baseline}. As expected, MLP is the worst performing baseline as it does not process the input as a sequence, which is vital to the problem statement. Vanilla LSTM is easily outperformed by adding attention mechanism to better handle longer sequences. However, the biggest jump in F1 score is seen in the performance of Multi-Scale LSTMs, which goes to show the significance of multi-context processing in such settings. Finally, our model outperforms the best performing baseline Multi-Scale LSTMs by 3.14\%, which emphasizes that our method of multi-context feature processing is superior to Multi-Scale LSTMs.

\begin{table}[h]
\centering
\begin{tabular}{|c|c|c|c|}
\hline
\textbf{Model} & \multicolumn{3}{|c|}{\textbf{Mean F1}} \\
\hline
 & \textbf{h=100} & \textbf{h=200} & \textbf{h=500} \\ 
\hline
MLP & 51.33\% & 50.97\% & 50.68\% \\
\hline
LSTM & 57.33\% & 55.27\% & 51.63\% \\
\hline
LSTM + Attention \cite{bahdanau2014neural} & 60.20\% & 59.58\% & 56.36\% \\
\hline
Multi-Scale LSTM \cite{liu2015multi} & 66.56\% & 63.98\% & 60.80\% \\
\hline
VLSTM & \textbf{69.70\%} & \textbf{68.41\%} & \textbf{65.90\%} \\
\hline
\end{tabular}
\caption{Comparison with other baseline models}
\label{tab:baseline}
\end{table}

\section{Challenges and Future Work}
\label{sec:conclusion}

LSTMs lead the field of time series analysis but are not equipped to deal with extremely long input sequences. We proposed a novel modification, VLSTMs, which can handle such extremely long sequences by simultaneously working at multiple levels. While our architecture is not restricted to just one domain, we do believe that it is designed to work in specific cases where gathering features from multiple contexts (or levels) makes sense. Our model contains an inherent assumption of periodicity for higher level feature extraction and thus might not translate well to situations where such an assumption can hurt the performance. For example, it might not be directly feasible for use in NLP, since it doesn't make any sense to read, say every 5th word in a document. However, the core idea of VLSTMs is to do multi-context feature processing independently and thus with further research, the same idea can be translated to these domains too.

\bibliographystyle{IEEEbib}
\bibliography{refs}

\begin{thebibliography}{10}

\bibitem{algo_trade}
Alain~P Chaboud, Benjamin Chiquoine, Erik Hjalmarsson, and Clara Vega,
\newblock ``Rise of the machines: Algorithmic trading in the foreign exchange
  market,''
\newblock {\em The Journal of Finance}, vol. 69, no. 5, pp. 2045--2084, 2014.

\bibitem{algo_trade2}
Terrence Hendershott, Ryan Riordan, et~al.,
\newblock ``Algorithmic trading and information,''
\newblock {\em Manuscript, University of California, Berkeley}, 2009.

\bibitem{RePEc:eee:jfinec:v:33:y:1993:i:1:p:3-5}
Eugene~F Fama and Kenneth~R French,
\newblock ``Common risk factors in the returns on stocks and bonds,''
\newblock {\em Journal of financial economics}, vol. 33, no. 1, pp. 3--56,
  1993.

\bibitem{hochreiter1997long}
Sepp Hochreiter and J{\"u}rgen Schmidhuber,
\newblock ``Long short-term memory,''
\newblock {\em Neural computation}, vol. 9, no. 8, pp. 1735--1780, 1997.

\bibitem{bahdanau2014neural}
Dzmitry Bahdanau, Kyunghyun Cho, and Yoshua Bengio,
\newblock ``Neural machine translation by jointly learning to align and
  translate,''
\newblock {\em arXiv preprint arXiv:1409.0473}, 2014.

\bibitem{Kearns_machinelearning}
Michael Kearns and Yuriy Nevmyvaka,
\newblock ``Machine learning for market microstructure and high frequency
  trading,''
\newblock {\em High Frequency Trading: New Realities for Traders, Markets, and
  Regulators}, 2013.

\bibitem{bid_ask}
``Bid and ask, mean reversion, investopedia,''
  https://www.investopedia.com/terms.

\bibitem{Chakraborty2011MarketMA}
Tanmoy Chakraborty and Michael Kearns,
\newblock ``Market making and mean reversion,''
\newblock in {\em Proceedings of the 12th ACM conference on Electronic
  commerce}. ACM, 2011, pp. 307--314.

\bibitem{dnns}
Andr{\'e}s Ar{\'e}valo, Jaime Ni{\~n}o, German Hern{\'a}ndez, and Javier
  Sandoval,
\newblock ``High-frequency trading strategy based on deep neural networks,''
\newblock in {\em International conference on intelligent computing}. Springer,
  2016, pp. 424--436.

\bibitem{book_algo}
Irene Aldridge,
\newblock {\em High-frequency trading: a practical guide to algorithmic
  strategies and trading systems}, vol. 604,
\newblock John Wiley \& Sons, 2013.

\bibitem{1709.01268}
Dat~Thanh Tran, Martin Magris, Juho Kanniainen, Moncef Gabbouj, and Alexandros
  Iosifidis,
\newblock ``Tensor representation in high-frequency financial data for price
  change prediction,''
\newblock in {\em 2017 IEEE Symposium Series on Computational Intelligence
  (SSCI)}. IEEE, 2017, pp. 1--7.

\bibitem{doi:10.1111/j.1540-6261.1991.tb02683.x}
Charles~MC Lee and Mark~J Ready,
\newblock ``Inferring trade direction from intraday data,''
\newblock {\em The Journal of Finance}, vol. 46, no. 2, pp. 733--746, 1991.

\bibitem{1011.6402}
Rama Cont, Arseniy Kukanov, and Sasha Stoikov,
\newblock ``The price impact of order book events,''
\newblock {\em Journal of financial econometrics}, vol. 12, no. 1, pp. 47--88,
  2014.

\bibitem{imbalance}
Darryl Shen,
\newblock {\em Order imbalance based strategy in high frequency trading},
\newblock Ph.D. thesis, oxford university, 2015.

\bibitem{1705.03233}
Adamantios Ntakaris, Martin Magris, Juho Kanniainen, Moncef Gabbouj, and
  Alexandros Iosifidis,
\newblock ``Benchmark dataset for mid-price prediction of limit order book
  data,''
\newblock {\em arXiv preprint arXiv:1705.03233}, 2017.

\bibitem{hlstm}
Junyoung Chung, Sungjin Ahn, and Yoshua Bengio,
\newblock ``Hierarchical multiscale recurrent neural networks,''
\newblock {\em arXiv preprint arXiv:1609.01704}, 2016.

\bibitem{rehlstm}
Akos K{\'a}d{\'a}r, Marc-Alexandre C{\^o}t{\'e}, Grzegorz Chrupa{\l}a, and Afra
  Alishahi,
\newblock ``Revisiting the hierarchical multiscale lstm,''
\newblock {\em arXiv preprint arXiv:1807.03595}, 2018.

\bibitem{liu2015multi}
Pengfei Liu, Xipeng Qiu, Xinchi Chen, Shiyu Wu, and Xuan-Jing Huang,
\newblock ``Multi-timescale long short-term memory neural network for modelling
  sentences and documents,''
\newblock in {\em Proceedings of the 2015 conference on empirical methods in
  natural language processing}, 2015, pp. 2326--2335.

\bibitem{wind}
Ignacio~A Araya, Carlos Valle, and H{\'e}ctor Allende,
\newblock ``Lstm-based multi-scale model for wind speed forecasting,''
\newblock in {\em Iberoamerican Congress on Pattern Recognition}. Springer,
  2018, pp. 38--45.

\bibitem{lu2018multi}
Rui Lu, Zhiyao Duan, and Changshui Zhang,
\newblock ``Multi-scale recurrent neural network for sound event detection,''
\newblock in {\em 2018 IEEE International Conference on Acoustics, Speech and
  Signal Processing (ICASSP)}. IEEE, 2018, pp. 131--135.

\bibitem{szegedy2015going}
Christian Szegedy, Wei Liu, Yangqing Jia, Pierre Sermanet, Scott Reed, Dragomir
  Anguelov, Dumitru Erhan, Vincent Vanhoucke, and Andrew Rabinovich,
\newblock ``Going deeper with convolutions,''
\newblock in {\em Proceedings of the IEEE conference on computer vision and
  pattern recognition}, 2015, pp. 1--9.

\bibitem{ntakaris2018benchmark}
Adamantios Ntakaris, Martin Magris, Juho Kanniainen, Moncef Gabbouj, and
  Alexandros Iosifidis,
\newblock ``Benchmark dataset for mid-price forecasting of limit order book
  data with machine learning methods,''
\newblock {\em Journal of Forecasting}, vol. 37, no. 8, pp. 852--866, 2018.

\bibitem{tsantekidis2017using}
Avraam Tsantekidis, Nikolaos Passalis, Anastasios Tefas, Juho Kanniainen,
  Moncef Gabbouj, and Alexandros Iosifidis,
\newblock ``Using deep learning to detect price change indications in financial
  markets,''
\newblock in {\em 2017 25th European Signal Processing Conference (EUSIPCO)}.
  IEEE, 2017, pp. 2511--2515.

\end{thebibliography}

\end{document}